\documentclass[conference]{IEEEtran}
\IEEEoverridecommandlockouts

\usepackage{amsmath,amssymb,amsfonts}
\usepackage{graphicx}
\usepackage{textcomp}
\usepackage{xcolor}
\usepackage{booktabs}
\usepackage{multirow}
\usepackage{tikz}
\usepackage{subcaption}
\usepackage[inline]{enumitem}
\usepackage[numbers,sort]{natbib}
\usepackage{nicefrac}
\usepackage{algorithm}
\usepackage[noend]{algpseudocode}
\usepackage{tabularx}
\usepackage[subpreambles=true]{standalone}
\usepackage{url}
\usepackage{listings}
\usepackage{float}
\usepackage{mystyle}

\usetikzlibrary{positioning, fit}

\def\BibTeX{{\rm B\kern-.05em{\sc i\kern-.025em b}\kern-.08em
    T\kern-.1667em\lower.7ex\hbox{E}\kern-.125emX}}

\usepackage{xspace}
\newcommand{\method}{Adaptive Safety through Knowledge\xspace}
\newcommand{\abbrev}{ASK\xspace}

\begin{document}

\title{%
    When to ASK: Uncertainty-Gated Language Assistance for Reinforcement Learning
}

\author{%
    \IEEEauthorblockN{Juarez Monteiro\textsuperscript{1}, Nathan Gavenski\textsuperscript{2}, Gianlucca Zuin\textsuperscript{1} and Adriano Veloso\textsuperscript{1}}
    \IEEEauthorblockA{%
        \begin{tabular}[t]{@{}c@{\qquad}c@{}}
            \textsuperscript{1}\textit{Kunumi Institute}        & \textsuperscript{2}\textit{King's College London} \\
            Belo Horizonte, MG, Brazil                                        & London, England, UK \\
            \{juarez, gianlucca, adriano\}@kunumi.com           & nathan.schneider\_gavenski@kcl.ac.uk
        \end{tabular}%
    }%
}

\maketitle

\begin{abstract}
Reinforcement learning (RL) agents often struggle with out-of-distribution (OOD) scenarios, leading to high uncertainty and random behavior. 
While language models (LMs) contain valuable world knowledge, larger ones incur high computational costs, hindering real-time use, and exhibit limitations in autonomous planning.
We introduce \method (\abbrev), which combines smaller LMs with trained RL policies to enhance OOD generalization without retraining.
\abbrev employs Monte Carlo Dropout to assess uncertainty and queries the LM for action suggestions only when uncertainty exceeds a set threshold. 
This selective use preserves the efficiency of existing policies while leveraging the language model's reasoning in uncertain situations. In experiments on the FrozenLake environment, \abbrev shows no improvement in-domain, but demonstrates robust navigation in transfer tasks, achieving a reward of $\mathbf{0.95}$.
Our findings indicate that effective neuro-symbolic integration requires careful orchestration rather than simple combination, highlighting the need for sufficient model scale and effective hybridization mechanisms for successful OOD generalization.
\end{abstract}


\section{Introduction\label{sec:introduction}}

Generalization is a vital requirement for learning systems, as it enables them to function beyond their training data.
In learning, these systems assume that the observed data are drawn from an independent and identically distributed distribution, and that this distribution holds approximately during evaluation and testing.
Yet, agentic applications do not exhibit this behavior.
During testing, even if the initial state is included in the training process, the sequential nature of episodes may push the agent out of distribution (OOD)~\cite{kirk2023survey}.

Generalizing to OOD states is challenging for agents because they lack prior information about how to act in these states.
When predicting in OOD situations, agents exhibit high \textit{uncertainty} and may behave randomly or be biased.
Thus, ensuring that agents behave appropriately in OOD states is also a vital component to develop safe and trusted systems.
Some works~\cite{garcia2015comprehensive,dalal2018safe,alshiekh2018safe} aim to ensure generalization by employing expert systems or incorporating prior knowledge to guide agent behavior in OOD states.
Recent advances in \textit{language models} (LM) offer a promising avenue: pre-trained on vast corpora, these models encode rich world knowledge and common-sense reasoning that can potentially guide agents when encountering OOD states.
However, integrating \textit{large} LM (LLMs) into real-time decision-making systems poses computational challenges.

While planners offer a safe alternative~\cite{mordoch2024safe,Galuppo2025Enhancing}, they require careful manual encoding of domain knowledge, limiting their adaptability to new tasks.
Smaller LMs, small to medium-sized models, offer an elegant trade-off: they retain much of the world knowledge and reasoning of larger models while being computationally efficient enough for real-time decision-making~\cite{chan2025retrieval}.
Unlike planners, LMs provide flexible guidance across diverse scenarios without requiring task-specific adjustments, leveraging existing trained policies to improve practical utility across different agents and domains.

Approaches to improving generalization can be broadly categorized into \textit{intrinsic} and \textit{extrinsic} methods.
Intrinsic methods, such as domain randomization, architectural innovations, or curriculum learning, modify the training process to encourage broader generalization, but require retraining agents with potentially larger networks or additional data~\cite{kirk2023survey}.
Extrinsic methods, conversely, augment existing policies at inference time without requiring retraining or additional training data.

In this work, we explore an extrinsic approach that integrates LMs with already-trained PPO agents to improve generalization in OOD settings, which we call \textit{\method} (\abbrev)\footnote{%
    \noindent\textbf{Data and Code Availability.}
    Supplementary material is available at \url{https://kclpure.kcl.ac.uk/ws/portalfiles/portal/368422362/appendix.pdf}.
    The code is available at \url{https://github.com/jrzmnt/ask-rl}.
}.
Our method leverages the semantic understanding and world knowledge encoded in LMs to guide reinforcement learning (RL) agents when they encounter unfamiliar states, combining the adaptability of RL with the implicit knowledge of language models without requiring retraining or architectural modifications.
\section{Problem Formulation\label{sec:problem}}

Reinforcement learning is a problem where a policy $\policy$ learns parameters $\agentW$ via interacting with the environment and its reward signal $r$~\cite{sutton2018reinforcement}.
In RL, the reward signal $r: S \times A \mapsto \mathbb{R}$, where $S$ is the state space and $A$ is the action space, allows the agent $\agent$ to learn the probability of each action $a \in A$ given the current state $s \in S$ by maximizing the immediate reward.
In this setting, agents typically do not observe the entire state space, resulting in a situation in which they lack state information and may perform highly uncertain actions.
To better formulate this change in unobserved states, we depart from the initial RL formulation and adopt one in which different contexts may not only alter how states are represented but also modify the transition functions that govern the environment's dynamics.

\subsection{Contextual Markov Decision Process\label{sec:sub:cmdp}}

We formulate our problem as a \textit{Contextual Markov Decision Process} (CMDP), a tuple $\mathcal{M} = \langle S, A, O, r, T, C, \phi \rangle$, where $O$ is the observation space, $r$ is the immediate reward function $r: O \times A \mapsto \mathbb{R}$, $T$ is the transition function $T: O \times A \mapsto O$, $C$ is the set of contexts, and $\phi$ is a projection function $\phi: S \times C \mapsto O$~\cite{kirk2023survey}.
Note that in CMDPs, the policy only observes the state produced by $\phi$, that the current context $c \in C$ remains the same in an episode, and the reward function $r$ now operates under observations and not states.
In other words, the policy perceives only one context $c$ at a time in which all states $s \in S$ are projected into it $\left\{\phi(s, c) \mid s \in S \right\}$.

In this formulation, the context serves as the seed, ID, or parameter vector that determines the current environment.
Hence, it should not change within an episode; it should change only between episodes.
More importantly, the context distribution $p(c)$ defines the training, evaluation, and testing collections of each task or environment instance.
Most often, this distribution remains uniform through all collections; however, this is not true for all environments.
Finally, we do not assume that the policy observes the context.
That is, the policy does not perceive its context.
Since two different contexts $(c, c')$ might map two different states $(s, s')$ into the same observation $(\phi(s, c) = \phi(s', c'))$, it is not always the case that the policy can acknowledge in which context it is in.

\subsection{Uncertainty in Neural Networks\label{sec:sub:uncertainty}}

Measuring uncertainty helps researchers assess the extent to which model predictions carry weight given their prior experience.
The two most common types of uncertainty in machine learning are:
\begin{enumerate*}[label=(\roman*)]
    \item \textit{epistemic}, which is inherent to models, caused by a lack of training data, hence reducible with more data; and
    \item \textit{aleatoric}, caused by inherent noise or ambiguity in data, thence irreducible with more data~\cite{kiureghian2009aleatory}.
\end{enumerate*}
Epistemic uncertainty at an observation $o \in O$ is high for a previously unseen $o$, and decreases as $o$ becomes part of the training set~\cite{kendall2017what}.
\citeauthor{kirsch2020unpacking}~\cite{kirsch2020unpacking} show that this behavior conforms with using mutual information in Bayesian models and deep ensembles, and \citeauthor{postels2020quantifying}~\cite{postels2020quantifying} as feature-space density in deterministic models as surrogates for epistemic uncertainty.
Aleatoric uncertainty at an observation $o \in O$ is high for ambiguous or noisy samples~\cite{kiureghian2009aleatory}.
For example, if multiple labels are observed at $o$, the aleatoric uncertainty will be high, thus, adding more data has no effect on it.

Uncertainty is usually measured under \textit{Bayesian models} or \textit{ensemble methods}~\cite{mukhoti2023deep}.
Bayesian models provide a principled way to quantify uncertainty by using prior distributions over model parameters to infer the posterior distribution given training data.
Ensemble models compute the average of their outputs, which are then used to estimate the entropy of the averaged vectors.
However, both approaches incur high training costs.
To represent uncertainty, these approaches double the number of parameters for the same network size and require more time to converge~\cite{gal2016droupout}.
Therefore, a solution for approximating the behavior of Bayesian models and ensemble methods is to apply dropout to the same neural network during inference $N$ times~\cite{gal2016droupout}.
By applying dropout during inference, we obtain multiple stochastic predictions from the same network, creating an implicit ensemble without additional parameters.
The variance across these predictions provides an estimate of epistemic uncertainty:
\begin{equation}\label{eq:epistemic}
    \textstyle \mathcal{U}_e(o) = \frac{1}{N} \sum_{i=1}^{N} \sum_{a} p_i(a \mid o) \log \frac{p_i(a \mid o)}{\bar{p}(a \mid o)},
\end{equation}
where $p_i(a \mid o)$ is the action distribution from the $i$-th forward pass with dropout, and $\bar{p}(a\mid o) = \nicefrac{1}{N} \sum_{i=1}^{N}p_i(a\mid o)$ is the mean distribution (we omit the agent notation $\agent$ from $p_i^\agent$ for simplicity).
On the other hand, the mean prediction entropy captures aleatoric uncertainty:
\begin{equation}\label{eq:aleatoric}
    \textstyle \mathcal{U}_a(o) = \frac{1}{N} \sum_{i=1}^{N} \sum_{a} -p_i(a \mid o) \log p_i(a \mid o).
\end{equation}
This Monte Carlo Dropout (MC Dropout) approach offers a lightweight alternative to full Bayesian methods or explicit ensembles, requiring only repeated forward passes through an already-trained network. 
To preserve the extrinsic nature and interpretability of ASK, our gating mechanism is implemented as a fixed uncertainty threshold, thus avoiding additional learned components or retraining.
In this setting, uncertainty is used as a trigger for intervention rather than as a calibrated decision signal, making precise calibration less critical.

\section{\method\label{sec:method}}
In this work, we propose \method (\abbrev), which leverages language models as an extrinsic method to improve the generalization of RL agents without any retraining.
\abbrev's main contribution lies in quantifying uncertainty to identify when an agent encounters OOD observations and selectively querying for guidance only in these high-uncertainty scenarios.
This approach combines the efficiency of a trained RL policy with the world knowledge of language models, activating it only when needed.
By measuring both epistemic and aleatoric uncertainty at each observation, \abbrev determines when the agent lacks sufficient experience (high epistemic uncertainty) or faces inherently ambiguous states (high aleatoric uncertainty).
In such cases, rather than allowing the agent to act with high uncertainty, we leverage the semantic understanding to provide an informed action recommendation.

We first summarize \abbrev.
Given a trained RL policy $\agent$, a language model $\generalist$, and a set of evaluation contexts $C' \in C$, where $C'$ are contexts not seen during training or during evaluation, \abbrev operates as follows.
For each context $c \in C'$, it initializes an environment instance and retrieves the initial observation $o$.
At each step, the agent computes an action $a$ using its learned policy $\agent(o)$.
However, before executing this action, \abbrev computes the total uncertainty $\mathcal{U}_e(o) + \mathcal{U}_a(o)$ using MC Dropout (Eqs.~\ref{eq:epistemic} and~\ref{eq:aleatoric}).
If the total uncertainty exceeds a predefined threshold $\threshold$, we override the agent's action by querying $\generalist$ with a prompt $Pr$ and the current observation to obtain an alternative action $\generalist(Pr, o)$.
This intervention occurs only when the agent is uncertain, allowing the trained policy to handle familiar observations while delegating high-uncertainty decisions to the language model's world knowledge.
The agent then executes the selected action, receives the next observation from the environment via the transition function $T$, and repeats this process until it reaches the goal state $g$.

\subsection{State Representation and Prompting\label{sec:sub:overview}}
Smaller language models do not share the same capacities as their larger counterparts.
As model size decreases, its ability to follow complex instructions, perform multi-step reasoning, and robustly handle varied prompt formats~\cite{chan2025retrieval} also decreases.
For example, when using LLMs, prompts that described a grid-like environment using a matrix would work perfectly.
Yet, when using smaller models, we had to simplify the observation description to better fit the model's capacity.
Therefore, careful prompt engineering is essential to maximize the utility of LMs in \abbrev.
\abbrev requires two key components: translating environmental observations into natural-language descriptions that models can interpret, and enforcing a structured output format that can be reliably parsed into discrete actions.

\abbrev prompt design balances informativeness with simplicity, providing the language model with just enough context to make informed decisions without overwhelming its reasoning capacity.
The complete prompt consists of several structured sections:
\begin{enumerate*}[label=(\roman*)]
    \item a role definition establishing the task and explicit output constraints to prevent verbose responses;
    \item high-level decision rules prioritizing safety and goal-directed behavior;
    \item state information using position coordinates and local spatial relationships; and
    \item the original policy's action suggestion as an ``autopilot recommendation.''
\end{enumerate*}
Appendix~\ref{ap:prompt} provides the full prompt.

Crucially, we include the trained policy's action recommendation in the prompt, framed as an ``autopilot suggestion.''
This allows the model to validate the policy's choice when it aligns with safe, goal-directed behavior, or override it when the suggestion appears risky given the current state.
By providing this suggestion, we leverage the RL policy's learned behavior while allowing the LM to intervene based on its knowledge.

Finally, \abbrev enforces a strict output format to facilitate reliable parsing and provides a fallback mechanism when the LM fails to adhere to the format instructions (by defaulting to the policy's action).
Importantly, \abbrev does \textit{not} fine-tune or train the language model.
It relies solely on the LM's pre-trained knowledge, maintaining its extrinsic nature.

\subsection{Uncertainty Estimation\label{sec:sub:uncertainty_estimation}}
\abbrev quantifies uncertainty using MC Dropout to obtain both epistemic and aleatoric estimates (Eqs.~\ref{eq:epistemic} and~\ref{eq:aleatoric}).
It computes the total uncertainty $\mathcal{U}_{\text{total}}(o) = \mathcal{U}_e(o) + \mathcal{U}_a(o)$ at each observation $o$ by performing $N$ forward passes through the trained policy network with dropout enabled.
This provides a lightweight mechanism for identifying states in which the agent lacks confidence in its action selection.

Uncertainty serves as a natural trigger for external reasoning in OOD states, rather than as a replacement for learned behavior.
High epistemic uncertainty indicates that the agent has insufficient experience with observations similar to $o$, precisely the scenario in which the agent is most likely to be out-of-distribution and where external guidance is most valuable.
High aleatoric uncertainty, on the other hand, signals inherent ambiguity in the observation itself, such as multiple valid paths or conflicting cues, where world knowledge and common-sense reasoning can help break ties.

By using uncertainty as the intervention criterion, \abbrev ensures that the LM is queried only when necessary.
In low-uncertainty states, the trained RL policy operates autonomously, leveraging its learned behavior efficiently.
In high-uncertainty states, where the agent exhibits high variance, the LM provides informed guidance based on its semantic understanding of navigation and safety.
This selective intervention preserves the efficiency of the learned policy while mitigating the risks associated with OOD generalization.

The threshold $\threshold$ provides a crucial trade-off between computational cost and safety.
Querying the LM at every step would be computationally expensive and unnecessary, as the trained policy performs well in familiar states.
Conversely, never querying the LM would leave the agent vulnerable to unsafe or random behavior in OOD scenarios.
By setting $\threshold$ empirically based on the uncertainty distribution observed during training, we balance policy autonomy with external guidance, invoking the LM only when the agent's uncertainty warrants the additional computational overhead.
\section{Experiments\label{sec:experiments}}

\begin{figure*}[t!hp]
    \centering
    \begin{minipage}{0.33\textwidth}
        \centering
        \begin{subfigure}{0.49\columnwidth}
            \centering
            \includestandalone[width=.75\linewidth]{figures/custom_trajectories/trajectories_5x5}
            \caption{\abbrev Difference}
            \label{fig:sub:same}
        \end{subfigure}
        \hfill
        \begin{subfigure}{0.49\columnwidth}
            \centering
            \includestandalone[width=.75\linewidth]{figures/custom_trajectories/trajectories_8x8}
            \caption{Different paths}
            \label{fig:sub:diff}
        \end{subfigure}
        \setcounter{figure}{0}
        \captionof{figure}{Different paths for the same context, where \textcolor{cb_purple}{purple} is the LM, \textcolor{cb_orange}{orange} the PPO, and \textcolor{cb_green}{green} the \abbrev approach.}
        \label{fig:paths}
    \end{minipage}
    \hfill
    \begin{minipage}{0.65\textwidth}
        \centering
        \includegraphics[width=.9\textwidth]{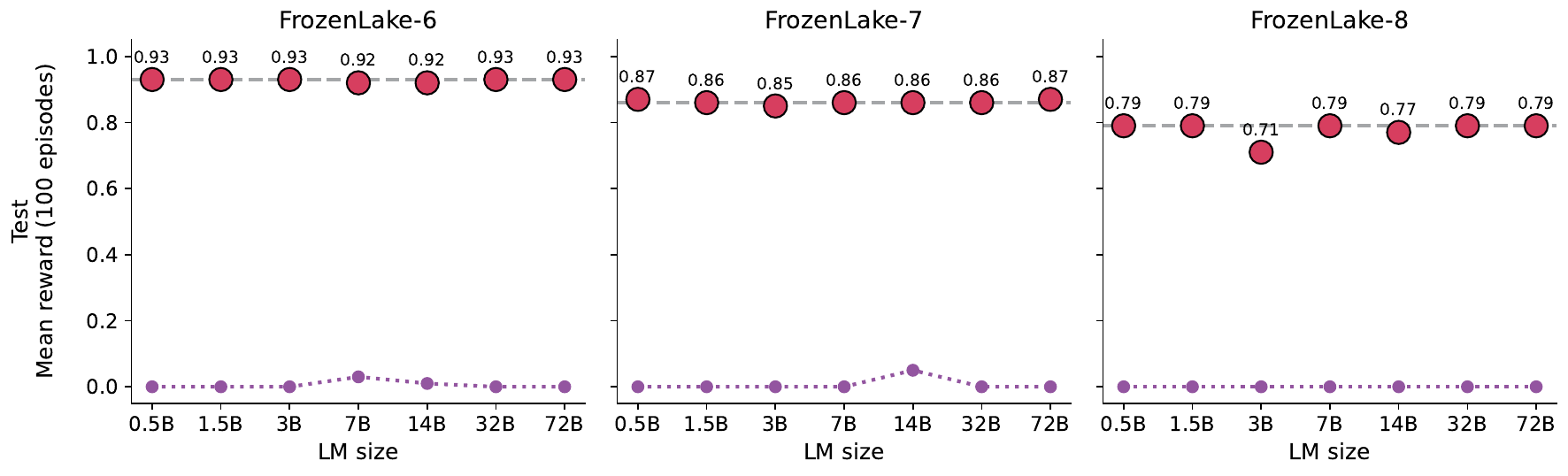}
        \begin{tikzpicture}
            \node[circle, fill={rgb,255:red,220;green,60;blue,100}, draw=black, minimum size=5pt, inner sep=0pt] (redcirc) at (0,0) {};
            \node[right=2pt of redcirc, font=\scriptsize] (redlabel) {\abbrev};

            \draw[color={rgb,255:red,147;green,85;blue,160}, dashed, thick] ([xshift=8pt]redlabel.east) -- ++(.3, 0) coordinate (lmdots);
            \node[circle, fill={rgb,255:red,147;green,85;blue,160}, minimum size=5pt, inner sep=0pt] (purpledot) at ([xshift=12.5pt]redlabel.east) {};
            \node[right=2pt of lmdots, font=\scriptsize] (purplelabel) {LM-only};

            \draw[gray, dashed, thick] ([xshift=8pt]purplelabel.east) -- ++(.3, 0) coordinate (ppodots);
            \node[right=2pt of ppodots, font=\scriptsize] (ppolabel) {PPO};
        \end{tikzpicture}
        \captionof{figure}{%
            In-domain performance of PPO, LM, and \abbrev on the FrozenLake environment across grid sizes 
            for language model sizes ranging from 0.5B to 72B parameters.
            Values above markers denote mean reward over 100 episodes.
        }
        \label{fig:domain_comparison}
    \end{minipage}
\end{figure*}


We evaluate \abbrev on the FrozenLake-v$1$ environment~\cite{towers2024gymnasium} across map sizes ranging from $4\!\times\!4$ to $8\!\times\!8$. 
For each map size, we create a set of $300$ contexts with different hole positions (as in Fig.~\ref{fig:paths}), which we equally split into training, evaluation, and testing sets.
By maintaining fixed contexts across splits, we ensure that no RL policy has access to complete information about contexts outside its training set.
Each configuration is evaluated over $100$ episodes, and results are reported as mean and standard deviation.
FrozenLake provides a simple yet informative environment for studying how uncertainty, learned policies, and language-based guidance interact. 
Varying map sizes and hole layouts enable the controlled induction of distribution shifts, thus reliably stressing learned policies and leading to familiar failure modes. 
This makes FrozenLake a useful diagnostic environment.
Our experiments aim to demonstrate that selectively invoking language guidance based on uncertainty can recover sensible behavior when both the RL policy and the language model fail on their own, a pattern expected to arise in more complex domains.

We adopt Proximal Policy Optimization (PPO)~\cite{ppo} as the reinforcement learning policy (from StableBaselines$3$~\cite{stablebaselines3}).
PPO is a widely used and well-established algorithm in the reinforcement learning literature, providing a strong and stable baseline for sequential decision-making.
We use fixed hyperparameters across all experiments and train each policy $2\!\times\!10^7$ timesteps, saving the best-performing model based on the evaluation set (see Appendix~\ref{ap:eval} for evaluation results).
For LMs, we select models from the Qwen family~\cite{qwen2.5}, using readily available weights from HuggingFace for reproducibility.
We employ these models off-the-shelf without any task-specific training or fine-tuning.
Using a single model family allows us to systematically vary model capacity while minimizing architectural and training-related confounding factors, ensuring that observed differences are mainly attributable to scale rather than changes in model design.
For the MC Dropout uncertainty estimation, we use $N=100$ forward passes with a dropout rate of $0.2$.

We evaluate each configuration by measuring average accumulated reward $\sum_i \gamma^i r_{i+1}$, where $\gamma$ is the discount factor.
Note that the FrozenLake reward function is sparse: the agent receives a reward of $1$ upon reaching the goal and $0$ otherwise~\cite{towers2024gymnasium, schmitt2025deontic}.
To assess efficiency, we also report the average episode length, capped at $100$ timesteps.
Episodes reaching maximum length with zero reward indicate the agent is stuck in a loop.
Although the original $8\!\times\!8$ implementation uses $200$ as maximum length, the trained PPO policies achieve the goal on $\approx12$ steps.
Thus, we standardize the maximum length to $100$ across all map sizes for a fair comparison.
Note that generalizing across contexts from maps of different sizes is more challenging than generalizing across maps of the same size with different hole positions.
Finally, we aim to assess the computational overhead and behavioral impact of LM integration.
To quantify this, we consider two metrics:

\noindent \textbf{Intervention rate (IR)}: measures the proportion of steps where uncertainty exceeds the threshold $\threshold$ and the LM is consulted:
$IR(\text{ep.}) = \nicefrac{1}{\vert\text{ep.}\vert} \sum_{t=1}^{\vert\text{ep.}\vert} 1\left[\mathcal{U}_e(o_t) + \mathcal{U}_a(o_t) \geqslant \threshold\right]$,
where $\vert\text{ep.}\vert$ is the episode length, $o_t$ is the observation at timestep $t$, and $1[\cdot]$ is the indicator function.
A low intervention rate indicates that the agent operates autonomously, while a high rate suggests frequent uncertainty requiring assistance.
The threshold $\tau$ was selected via Bayesian optimization with Optuna~\cite{akiba2019optuna} 
over $[0.10, 1.20]$ (22 independent studies).
Mid-size models (3B--14B) require high thresholds ($\tau > 0.9$) to suppress unreliable interventions, while 1.5B and 72B operate at lower values ($\tau < 0.55$), reflecting better-calibrated guidance.
Notably, 72B achieves $\tau = 0.12$ on $7{\times}7$, consulting the LM at nearly 
every step while maintaining 0.89 reward --- a behavior only sustainable at sufficient 
model scale.

\noindent \textbf{Overwrite rate (OR)}: measures how often the recommended action differs from the original policy's suggestion when consulted:
$OR(\text{ep.}) = \nicefrac{\sum_{t=1}^{\vert\text{ep.}\vert} 1\left[\mathcal{U}_e(o_t) \!+\! \mathcal{U}_a(o_t) \geqslant \threshold \land a_t^{\generalist} \neq a_t^{\agent}\right]}{IR(\text{ep.})}$,
where $a_t^{\generalist}$ is the action selected by the LM and $a_t^{\agent}$ is the action suggested by the trained policy.
This metric reveals whether the LM merely validates the policy's choices or provides genuinely different guidance.
A high overwrite rate indicates that the LM frequently disagrees with the policy in uncertain states, suggesting it offers complementary reasoning rather than redundant confirmation.

Combined, these metrics provide an efficiency-performance trade-off insight. IR quantifies computational overhead, while OR assesses the added value of LM guidance when invoked.

\subsection{In-Domain Evaluation (Same-Size Maps)\label{sec:sub:same}}

\begin{table*}[t]
\scriptsize
\centering
\caption{Comparison of PPO baseline and \abbrev on FrozenLake: same-size evaluation and downward generalization.}
\label{tab:combined_results}
\setlength{\belowrulesep}{0.5pt}
\setlength{\aboverulesep}{0.5pt}
\begin{tabularx}{\textwidth}{ll *{4}{>{\centering\arraybackslash}X} *{4}{>{\centering\arraybackslash}X}}
\toprule
& & \multicolumn{4}{c}{(a) Same-size evaluation} & \multicolumn{4}{c}{(b) Downward generalization (trained on 8×8)} \\[-2pt]
\cmidrule(lr){3-6} \cmidrule(lr){7-10}
Size & Model & Reward & Length & IR (\%) & OR (\%) & Reward & Length & IR (\%) & OR (\%) \\[-2pt]
\midrule

\multirow{8}{*}{$4 \times 4$}
 & PPO  & -- & -- & -- & -- & $0.00 \pm 0.00$ & $46.98 \pm 49.18$ & -- & -- \\[-1pt]
 & 0.5B & -- & -- & -- & -- & $0.00 \pm 0.00$ & $41.12 \pm 26.20$ & 100.00 & 61.71 \\[-1pt]
 & 1.5B & -- & -- & -- & -- & $0.37 \pm 0.49$ & $6.92 \pm 6.20$ & 100.00 & 42.98 \\[-1pt]
 & 3B   & -- & -- & -- & -- & $0.00 \pm 0.00$ & $44.58 \pm 26.47$ & 100.00 & 64.52 \\[-1pt]
 & 7B   & -- & -- & -- & -- & $0.00 \pm 0.00$ & $61.00 \pm 0.00$ & 100.00 & 95.57 \\[-1pt]
 & 14B  & -- & -- & -- & -- & $0.07 \pm 0.26$ & $56.59 \pm 15.04$ & 100.00 & 52.95 \\[-1pt]
 & 32B  & -- & -- & -- & -- & $\mathbf{0.95 \pm 0.22}$ & $\mathbf{7.70 \pm 9.46}$ & 100.00 & 46.17 \\[-1pt]
 & 72B  & -- & -- & -- & -- & $0.95 \pm 0.22$ & $7.99 \pm 9.51$ & 100.00 & 64.22 \\[-2pt]
\midrule

\multirow{8}{*}{$5 \times 5$}
 & PPO  & -- & -- & -- & -- & $0.010 \pm 0.099$ & $42.41 \pm 48.26$ & -- & -- \\[-1pt]
 & 0.5B & -- & -- & -- & -- & $0.00 \pm 0.00$ & $31.61 \pm 27.21$ & 100.00 & 99.87 \\[-1pt]
 & 1.5B & -- & -- & -- & -- & $0.23 \pm 0.42$ & $11.53 \pm 13.13$ & 100.00 & 28.47 \\[-1pt]
 & 3B   & -- & -- & -- & -- & $0.00 \pm 0.00$ & $50.92 \pm 22.39$ & 100.00 & 41.07 \\[-1pt]
 & 7B   & -- & -- & -- & -- & $0.00 \pm 0.00$ & $61.00 \pm 0.00$ & 100.00 & 51.03 \\[-1pt]
 & 14B  & -- & -- & -- & -- & $0.01 \pm 0.10$ & $58.11 \pm 12.70$ & 100.00 & 47.27 \\[-1pt]
 & 32B  & -- & -- & -- & -- & $\mathbf{0.87 \pm 0.34}$ & $\mathbf{10.12 \pm 10.71}$ & 100.00 & 47.55 \\[-1pt]
 & 72B  & -- & -- & -- & -- & $0.86 \pm 0.35$ & $16.18 \pm 18.42$ & 100.00 & 50.51 \\[-2pt]
\midrule

\multirow{8}{*}{$6\times6$}
 & PPO  & $0.93 \pm 0.26$ & $\mathbf{9.49 \pm 1.90}$ & -- & -- & $0.00 \pm 0.00$ & $36.80 \pm 46.62$ & -- & -- \\[-1pt]
 & 0.5B & $\mathbf{0.93 \pm 0.26}$ & $\mathbf{9.49 \pm 1.90}$ & 49.83 & 0.00 & $0.00 \pm 0.00$ & $31.61 \pm 25.91$ & 100.00 & 100.00 \\[-1pt]
 & 1.5B & $\mathbf{0.93 \pm 0.26}$ & $\mathbf{9.49 \pm 1.90}$ & 55.03 & 0.00 & $0.11 \pm 0.31$ & $8.60 \pm 8.27$ & 100.00 & 28.96 \\[-1pt]
 & 3B   & $\mathbf{0.93 \pm 0.26}$ & $\mathbf{9.49 \pm 1.90}$ & 0.00 & 0.00 & $0.00 \pm 0.00$ & $50.38 \pm 22.78$ & 100.00 & 41.43 \\[-1pt]
 & 7B   & $0.92 \pm 0.27$ & $11.66 \pm 12.95$ & 27.19 & 4.09 & $0.00 \pm 0.00$ & $61.00 \pm 0.00$ & 100.00 & 50.31 \\[-1pt]
 & 14B  & $0.92 \pm 0.27$ & $9.55 \pm 2.13$ & 0.97 & 0.47 & $0.01 \pm 0.10$ & $60.31 \pm 5.17$ & 100.00 & 51.50 \\[-1pt]
 & 32B  & $\mathbf{0.93 \pm 0.26}$ & $\mathbf{9.49 \pm 1.90}$ & 32.03 & 0.00 & $0.69 \pm 0.46$ & $22.92 \pm 21.88$ & 100.00 & 52.35 \\[-1pt]
 & 72B  & $\mathbf{0.93 \pm 0.26}$ & $9.50 \pm 1.86$ & 24.03 & 0.60 & $\mathbf{0.75 \pm 0.44}$ & $\mathbf{21.58 \pm 21.09}$ & 100.00 & 50.33 \\[-2pt]
\midrule

\multirow{8}{*}{$7\times7$}
 & PPO  & $0.86 \pm 0.35$ & $13.08 \pm 12.58$ & -- & -- & $0.00 \pm 0.00$ & $30.22 \pm 43.76$ & -- & -- \\[-1pt]
 & 0.5B & $\mathbf{0.87 \pm 0.34}$ & $\mathbf{11.45 \pm 2.46}$ & 58.23 & 0.26 & $0.00 \pm 0.00$ & $25.28 \pm 24.98$ & 100.00 & 99.95 \\[-1pt]
 & 1.5B & $0.86 \pm 0.35$ & $13.08 \pm 12.64$ & 41.59 & 0.00 & $0.10 \pm 0.30$ & $8.08 \pm 7.06$ & 100.00 & 29.14 \\[-1pt]
 & 3B   & $0.85 \pm 0.36$ & $13.03 \pm 12.65$ & 0.56 & 0.14 & $0.00 \pm 0.00$ & $53.51 \pm 19.53$ & 100.00 & 44.23 \\[-1pt]
 & 7B   & $0.86 \pm 0.35$ & $14.25 \pm 15.27$ & 9.31 & 2.53 & $0.00 \pm 0.00$ & $61.00 \pm 0.00$ & 100.00 & 51.57 \\[-1pt]
 & 14B  & $0.86 \pm 0.35$ & $13.08 \pm 12.64$ & 0.00 & 0.00 & $0.00 \pm 0.00$ & $61.00 \pm 0.00$ & 100.00 & 50.51 \\[-1pt]
 & 32B  & $0.86 \pm 0.35$ & $13.08 \pm 12.64$ & 56.62 & 0.00 & $0.58 \pm 0.50$ & $30.20 \pm 23.38$ & 100.00 & 53.36 \\[-1pt]
 & 72B  & $\mathbf{0.87 \pm 0.34}$ & $12.22 \pm 9.09$ & 58.32 & 0.32 & $\mathbf{0.68 \pm 0.47}$ & $\mathbf{27.10 \pm 21.94}$ & 100.00 & 50.88 \\[-2pt]
\midrule

\multirow{8}{*}{$8\times8$}
 & PPO  & $\mathbf{0.79 \pm 0.41}$ & $\mathbf{12.49 \pm 3.17}$ & -- & -- & -- & -- & -- & -- \\[-1pt]
 & 0.5B & $\mathbf{0.79 \pm 0.41}$ & $12.52 \pm 3.21$ & 59.18 & 0.13 & -- & -- & -- & -- \\[-1pt]
 & 1.5B & $\mathbf{0.79 \pm 0.41}$ & $\mathbf{12.49 \pm 3.19}$ & 36.61 & 0.00 & -- & -- & -- & -- \\[-1pt]
 & 3B   & $0.71 \pm 0.46$ & $18.14 \pm 21.14$ & 14.74 & 6.63 & -- & -- & -- & -- \\[-1pt]
 & 7B   & $\mathbf{0.79 \pm 0.41}$ & $15.49 \pm 13.93$ & 14.99 & 2.88 & -- & -- & -- & -- \\[-1pt]
 & 14B  & $0.77 \pm 0.42$ & $15.10 \pm 15.10$ & 4.32 & 3.20 & -- & -- & -- & -- \\[-1pt]
 & 32B  & $\mathbf{0.79 \pm 0.41}$ & $\mathbf{12.49 \pm 3.19}$ & 43.01 & 0.00 & -- & -- & -- & -- \\[-1pt]
 & 72B  & $\mathbf{0.79 \pm 0.41}$ & $12.50 \pm 3.17$ & 28.60 & 0.54 & -- & -- & -- & -- \\[-2pt]
\bottomrule
\end{tabularx}
\vspace*{-0.3cm}
\end{table*}

Contrary to our expectations, Table~\ref{tab:combined_results}a shows that integrating LMs does not consistently improve over the PPO baseline.
Across all three environment sizes, most LM-augmented models achieve reward values statistically indistinguishable from the PPO baseline.
On the $6\!\times\!6$ grid, PPO achieves $0.93 \pm 0.26$ reward with an episode length of $9.49 \pm 1.90$ steps.
While several models ($0.5$B, $1.5$B, $3$B, $32$B, $72$B) match this reward, they do so without meaningfully reducing episode length, and mid-size models ($7$B, $14$B) show slight performance degradation ($0.92 \pm 0.27$).
This pattern persists across larger grids: on $8\!\times\!8$, where the navigation challenge is most pronounced, only the smallest models ($0.5$B, $1.5$B) and the largest model ($72$B) maintain the baseline reward of $0.79 \pm 0.41$, while $3$B shows a substantial drop to $0.71 \pm 0.46$.

To understand the role of the uncertainty mechanism, we also evaluated an LM-only baseline where the language model directly controls the agent without PPO guidance.
Fig.~\ref{fig:domain_comparison} reveals that LMs alone fundamentally cannot solve these navigation tasks: on test splits, LM-only performance remains near zero across all model sizes and grid dimensions, and on evaluation sets, performance reaches at most $0.60$--$0.80$ on the simplest $6\!\times\!6$ grid, degrading to $0.60$--$0.70$ on $8\!\times\!8$.
This confirms our discussion in the related work (Section~\ref{sec:relatedwork}) regarding the limited planning capabilities of language models in sequential decision-making tasks.
Despite their strong performance on language understanding and generation, LMs lack the look-ahead search and value estimation mechanisms inherent to RL algorithms.
The near-zero validation performance is particularly striking, even with explicit environmental descriptions and step-by-step reasoning prompts, the models fail to navigate effectively in distribution shifts.
This underscores the necessity of the hybrid architecture: while LMs do not improve upon PPO in the same-size scenarios, the gating mechanism is essential to prevent the catastrophic performance degradation that would result from naive LM control.

The failure to improve performance appears closely related to the LM overwrite behavior.
Models that maintain baseline performance exhibit near-zero OR ($\leq 0.6\%$), suggesting they primarily defer to the PPO policy.
Conversely, models with higher OR, particularly the $3$B model with $6.63\%$ overwrites on $8\!\times\!8$, show marked performance degradation.
The episode length metric further corroborates this: when $3$B overwrites PPO's actions, episodes increase to $18.14 \pm 21.14$ steps, compared with the baseline's $12.49 \pm 3.17$, indicating that LM interventions often lead to suboptimal trajectories.

An interesting pattern emerges when comparing the smallest and largest models.
The $0.5$B model demonstrates high LM engagement ($49.83\%$ to $59.18\%$ calls across grid sizes) paired with minimal overwrites ($0.00\%$ to $0.26\%$), suggesting that its uncertainty estimates align well with PPO's competence—it queries frequently but rarely disagrees.
In contrast, the $72$B model exhibits comparable performance with more moderate call rates ($24.03\%$ to $58.32\%$) but slightly higher OR ($0.32\%$ to $0.60\%$).
This indicates that while $72$B is more selective about when to consult the LM, it is also more willing to override PPO when it does, though still with sufficient accuracy to maintain baseline performance.

Most surprisingly, mid-size models ($3$B, $7$B, $14$B) exhibit a failure mode characterized by low call rates coupled with high overwrite percentages.
For instance, $14$B on $8\!\times\!8$ calls the LM only $4.32\%$ of the time but overwrites $3.20\%$ of those calls, suggesting poor calibration between uncertainty estimation and actual LM reliability.
This U-shaped performance curve, where very small and very large models succeed while mid-size models struggle, was unexpected and suggests that the gating effectiveness may be size dependent.
These results led us to question whether the same-size evaluation was sufficiently challenging to reveal the potential benefits of LM guidance.
In-domain scenarios may be too simple, allowing PPO alone to already find near-optimal policies, leaving little room for LM-based improvement.
To investigate whether LMs could provide value in more demanding generalization scenarios, we next examine performance on different-size maps.

\subsection{Downward Generalization (Different-Size Maps)\label{sec:sub:different}}

Having established that same-size generalization provides insufficient challenge, we examine downward generalization where agents trained on $8\!\times\!8$ grids navigate smaller environments ($4\!\times\!4$ through $7\!\times\!7$).
Table~\ref{tab:combined_results}b shows that while neither component succeeds in isolation, their combination unlocks robust generalization at scale.

The PPO baseline fails to generalize downward, achieving zero reward as episode length remain closely the same from $46.98$ steps on $4\!\times\!4$ to $30.22$ steps on $7\!\times\!7$.
Combined with our earlier finding that LMs alone achieve near-zero performance (Fig.~\ref{fig:domain_comparison}).
A sharp capability threshold emerges at $32$B parameters.
Both $32$B and $72$B models achieve exceptional performance: $0.95 \pm 0.22$ reward on $4\!\times\!4$ with efficient navigation ($7.70$ and $7.99$ steps).
This success persists across all sizes: $0.86$--$0.87$ on $5\!\times\!5$, $0.69$--$0.75$ on $6\!\times\!6$, and $0.58$--$0.68$ on $7\!\times\!7$ where all other configurations fail completely.
Performance degrades with environment size, yet remains functional throughout.
This performance transition at $32$B reflects a capability threshold for reliable spatial reasoning under distribution shift, rather than a requirement imposed by ASK itself.
ASK is model-agnostic and exposes this trade-off by invoking language guidance only when policy uncertainty warrants intervention.

The relationship between model size and generalization exhibits a non-uniform pattern.
The $1.5$B model shows modest success ($0.37$ on $4\!\times\!4$, $0.23$ on $5\!\times\!5$) with short episodes.
However, mid-size models ($3$B--$14$B) fail completely, achieving zero reward despite often reaching maximum episode length.
This behavior, in which $1.5$B outperforms $3$B--$14$B but falls short of $32$B/$72$B, suggests a capability phase transition.

LM overwrite patterns reveal nuanced dynamics.
The $0.5$B model's very high overwrite ($61$--$100\%$) indicates overconfident interventions that consistently degrade performance.
The $1.5$B model maintains low overwrite ($28$--$43\%$), suggesting well-calibrated deference to PPO when uncertain.
Successful models ($32$B/$72$B) operate with moderate overwrite ($46$--$68\%$), indicating that the LM actively contributes by overriding PPO approximately half the time, and that these interventions are beneficial.
Overwrite rates exceeding $80\%$ indicate failure, where LM overconfidence leads to poor trajectories (Fig.~\ref{fig:domain_comparison}).

These results show that PPO alone yields zero reward, and LMs alone fail fundamentally.
Yet, their integration at scale ($32$B+) unlocks robust generalization.
This capability threshold suggests that downward generalization requires sophisticated spatial reasoning and goal-directed planning—capabilities that emerge in LMs only at scale, yet require RL's grounding to translate into effective navigation.
Most encouragingly, $32$B/$72$B models maintain functional performance, demonstrating genuinely transferable navigation strategies.
The synergy between LM guidance and RL optimization, which is hidden in same-size scenarios due to task simplicity, becomes evident under larger distribution shifts.
\section{Related Work\label{sec:relatedwork}}

Recent work has explored integrating language models into RL and planning tasks.
Several studies demonstrate that LLMs can generate plans when provided with appropriate context~\cite{huang2022language}, yet they face significant limitations in sequential decision-making.
\citeauthor{armony2025far}~\cite{armony2025far} show that even state-of-the-art LLMs struggle to match classical symbolic planners in structured domains, particularly when planning requires precise state tracking and multi-step reasoning.
Similarly, \citeauthor{valmeekam2023planning}~\cite{valmeekam2023planning} demonstrate that LLMs often fail to maintain consistency in long-horizon tasks and struggle with domains requiring exact logical reasoning.
These limitations motivate our approach: rather than using language models as planners, \abbrev queries LMs only in uncertain states where reasoning and world knowledge can complement learned policies.

Uncertainty quantification has been proposed as a mechanism for safe RL~\cite{kendall2017what}.
Monte Carlo Dropout~\cite{gal2016droupout} provides a computationally efficient method to estimate both epistemic and aleatoric uncertainty in neural networks by treating dropout as approximate Bayesian inference.
While prior work uses uncertainty to trigger human oversight or conservative actions, we instead use it to gate language model queries, providing automated guidance without human supervision or the computational overhead of full Bayesian inference.

\abbrev differs from existing LLM-RL integration approaches in three key ways:
\begin{enumerate*}[label=(\roman*)]
    \item usage of small to medium language models rather than large ones, making real-time deployment feasible;
    \item selectively invoking the LM based on uncertainty rather than at every step; and
    \item treating LMs as a fallback mechanism for OOD states rather than as the primary decision-maker.
\end{enumerate*}
This design acknowledges both the planning limitations of language models and the value of learned RL policies, using each where it excels.
\section{Conclusion\label{sec:conclusion}}

In this work, we presented \method (\abbrev), an extrinsic approach that improves OOD generalization in RL by selectively integrating LMs based on uncertainty estimates.
Using Monte Carlo Dropout, \abbrev identifies unfamiliar states and queries an LM only when policy uncertainty exceeds a predefined threshold, preserving efficiency while leveraging language-based commonsense reasoning in critical situations.
Empirically, while this integration yields no gains in-domain where PPO already performs well, it enables robust downward transfer: in scenarios where both PPO and language models alone fail, their uncertainty-gated combination achieves up to $0.95$ reward, demonstrating a strong synergy that emerges only at sufficient model scale.

More broadly, our results suggest that effective neuro-symbolic systems require principled orchestration rather than naive integration.
Instead of replacing specialist policies with generalist models, \abbrev demonstrates that combining RL for efficient decision-making with language models for reasoning under uncertainty yields more reliable and adaptable agents.
Our focus is not to outperform specialized safety or planning heuristics, but to demonstrate a general, domain-agnostic integration principle that augments existing policies without retraining or task-specific engineering.
In \abbrev, language models are not treated as planners, but as semantic critics that selectively override high-uncertainty actions, consistent with the moderate overwrite rates observed in successful configurations. The central question, therefore, is not whether language models can plan or RL agents can generalize, but when and how their complementary strengths should be combined.

Future work includes learning adaptive gating mechanisms beyond fixed thresholds, improving uncertainty calibration through ensembles or Bayesian methods, and exploring joint training of the policy and gate. 
Extending our approach to more complex and partially observable environments is a natural next step toward understanding the generality and robustness of uncertainty-gated language guidance.

\section*{Acknowledgment}
This work was partially supported by UK Research and Innovation [grant number EP/S023356/1], in the UKRI Centre for Doctoral Training in Safe and Trusted Artificial Intelligence (\url{www.safeandtrustedai.org}), and by the Kunumi Institute, through individual grants awarded to the authors.

\bibliographystyle{IEEEtranN}
\bibliography{reference}

\clearpage
\appendix[Additional Experimental Details]

\section{Prompt} \label{ap:prompt}

The prompt $Pr$ provides the language model with a structured description of the local navigation problem, encoding both the agent state and its surrounding environment.
All variables are deterministically extracted from the FrozenLake grid and injected into the prompt during inference.

The variables $\texttt{agent\_row}$ and $\texttt{agent\_col}$ (Line~\ref{code:agent}) denote the current row and column of the agent, while \texttt{goal\_row} and \texttt{goal\_col} (Line~\ref{code:goal}) represent the coordinates of the goal state.
Together, these variables provide global positional context, allowing the model to reason about relative distance and direction toward the goal.

The immediate surroundings of the agent are described by the variables \texttt{up\_tile}, \texttt{down\_tile}, \texttt{left\_tile}, and \texttt{right\_tile} (Lines~\ref{code:up_tile}--\ref{code:right_tile}), which encode the tile types directly adjacent to the agent in each cardinal direction.
This information enables the model to assess the immediate safety of each possible action, including the presence of holes or map boundaries.

To incorporate short-horizon foresight, the prompt additionally includes the look-ahead variables \path{up_up_tile}, \path{down_down_tile}, \path{left_left_tile}, and \path{right_right_tile} (Lines~\ref{code:up_up_tile}--\ref{code:right_right_tile}), representing the tiles located two steps away in the corresponding directions.
These variables allow the model to anticipate unsafe transitions beyond the immediate move and avoid actions that lead to future dead ends.

The action space is explicitly constrained to the discrete set $\{\texttt{UP}, \texttt{DOWN}, \texttt{LEFT}, \texttt{RIGHT}\}$ (Line~\ref{code:actions}), and the prompt enforces strict output rules requiring the model to return exactly one valid action without additional text or explanations.
Edge tiles are treated as unsafe, and when multiple safe actions are available, the model is instructed to select the action that minimizes the distance to the goal.

This structured prompt design enables deterministic, safety-aware reasoning by the language model while ensuring full compatibility with the discrete action space and transition dynamics of the reinforcement learning environment.

\begin{lstlisting}[
  basicstyle=\ttfamily\tiny,
  breaklines=true,
  numbers=left,
  numbersep=-6pt,
  escapechar=|,
  float=h,
  frame=lines,
  captionpos=t,
  caption={Prompt for \abbrev},
  label=lst:mycode
]
    You are a robot navigation policy.
    Your task is to choose exactly ONE action.

    VALID ACTIONS:
    UP
    DOWN
    LEFT
    RIGHT

    RULES:
    - Do NOT explain.
    - Do NOT add text.
    - Do NOT add markdown.
    - Do NOT add JSON keys other than "action".
    - EDGE tiles are unsafe and must not be chosen.
    - If the suggested action is unsafe, choose a safe alternative.
    - If multiple safe actions exist, choose the one that moves closer to the goal.

    STATE:
    Agent position: row {agent_row}, col {agent_col} |\label{code:agent}|
    Goal position: row {goal_row}, col {goal_col} |\label{code:goal}|

    IMMEDIATE NEIGHBORS:
    UP: {up_tile} |\label{code:up_tile}|
    DOWN: {down_tile} |\label{code:down_tile}|
    LEFT: {left_tile} |\label{code:left_tile}|
    RIGHT: {right_tile} |\label{code:right_tile}|

    LOOK AHEAD:
    UP->UP: {up_up_tile} |\label{code:up_up_tile}|
    DOWN->DOWN: {down_down_tile} |\label{code:down_down_tile}|
    LEFT->LEFT: {left_left_tile} |\label{code:left_left_tile}|
    RIGHT->RIGHT: {right_right_tile} |\label{code:right_right_tile}|

    OUTPUT FORMAT (MANDATORY):
    {"action":"UP"} OR {"action":"DOWN"} OR {"action":"LEFT"} OR {"action":"RIGHT"} |\label{code:actions}|
\end{lstlisting}

\section{FrozenLake Environment}
FrozenLake~\cite{towers2024gymnasium} is a discrete grid-world navigation environment commonly used as a benchmark for reinforcement learning algorithms.
The environment consists of a two-dimensional grid of size $N \times N$, where each cell represents a Markovian state and the agent occupies exactly one cell at each time step.

Each grid cell is associated with a semantic tile type.
The start tile (\textbf{S}) defines the initial state of the agent at the beginning of an episode, while the goal tile (\texttt{$g$}) represents the terminal success state.
Hole tiles (\textbf{H}) correspond to terminal failure states, immediately ending the episode upon entry.
All remaining cells correspond to safe frozen tiles on which the agent can visit without immediate termination.

The environment supports both deterministic and stochastic transition dynamics.
In the stochastic variant, actions may lead to unintended transitions due to slippage on the ice.
In this work, we exclusively adopt the deterministic formulation, in which each action deterministically moves the agent to the adjacent cell in the selected direction, unless the action would cause the agent to leave the grid boundaries, in which case the agent remains in its current position.

At each step, the agent selects one action from the discrete action space $\{\texttt{UP}, \texttt{DOWN}, \texttt{LEFT}, \texttt{RIGHT}\}$.
An episode terminates when the agent reaches either the goal tile or any hole tile.
A reward of $+1$ is assigned upon reaching the goal state, while all other transitions, including those leading to failure, yield a reward of $0$.
No intermediate shaping rewards are provided, making the task sparse-reward and requiring effective exploration and planning.

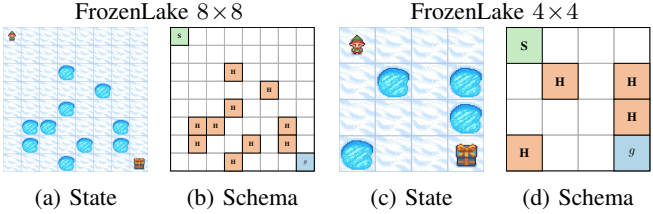
\begin{figure}[h!tp]
    \centering

    \begin{subfigure}{0.49\columnwidth}
        \caption*{FrozenLake $8\!\times\!8$}
        \begin{subfigure}{0.49\linewidth}
            \centering
            \includegraphics[height=.9\linewidth]{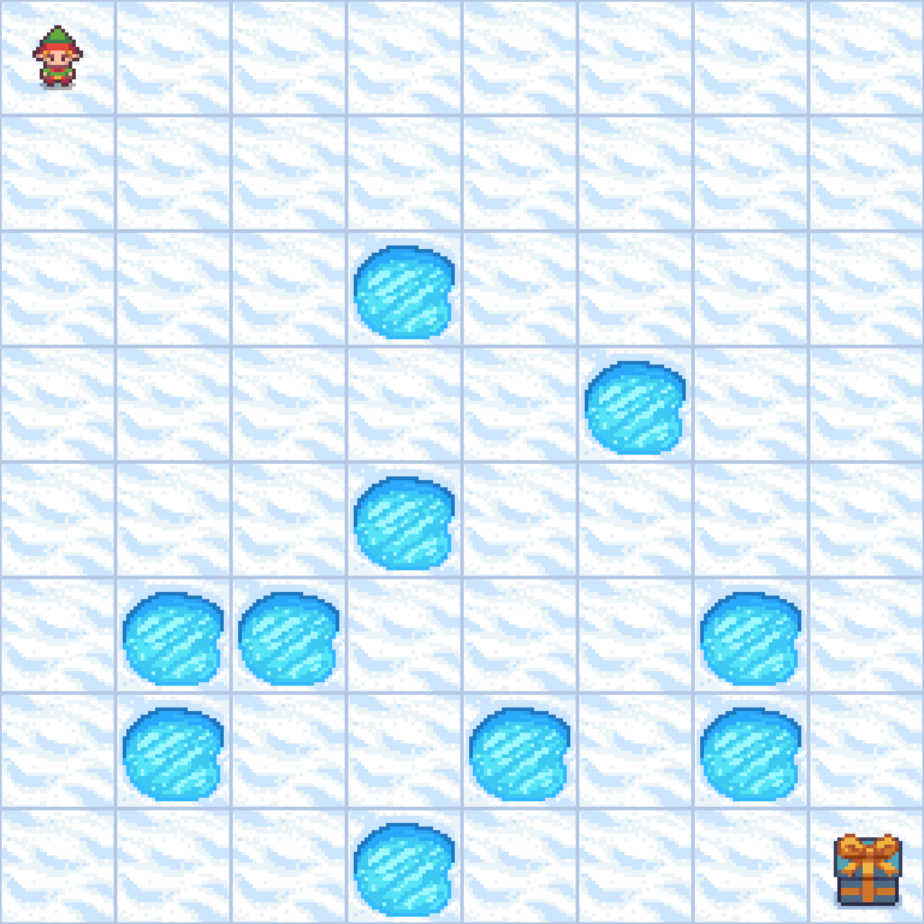}
            \caption{State}
            \label{fig:sub:gym8}
        \end{subfigure}
        \hfill
        \begin{subfigure}{0.49\linewidth}
              \centering
                  \begin{tikzpicture}
    
    \foreach \x in {0,...,7} {
      \foreach \y in {0,...,7} {
        \draw[gray!60] (\x,\y) rectangle ++(1,1);
      }
    }

    \draw[black] (0, 0) rectangle ++(8, 8);
    
    \filldraw[fill=cb_green!30] (0,7) rectangle ++(1,1);
    \node at (0.5,7.5) {\textbf{S}};
    
    \filldraw[fill=cb_blue!30] (7,0) rectangle ++(1,1);
    \node at (7.5,0.5) {$g$};
    
    \foreach \x/\y in {
      3/5,
      5/4,
      3/3,
      6/2,
      6/1,
      1/1,
      1/2,
      2/2,
      3/0,
      4/1
    } {
      \filldraw[fill=cb_orange!40] (\x,\y) rectangle ++(1,1);
      \node at (\x+0.5,\y+0.5) {\textbf{H}};
    }
    
    \end{tikzpicture}
              \caption{Schema}
              \label{fig:sub:schem8}
        \end{subfigure}
    \end{subfigure}
    \begin{subfigure}{0.49\columnwidth}
        \caption*{FrozenLake $4\!\times\!4$}
        \begin{subfigure}{0.49\linewidth}
            \centering
            \includegraphics[height=.9\textwidth]{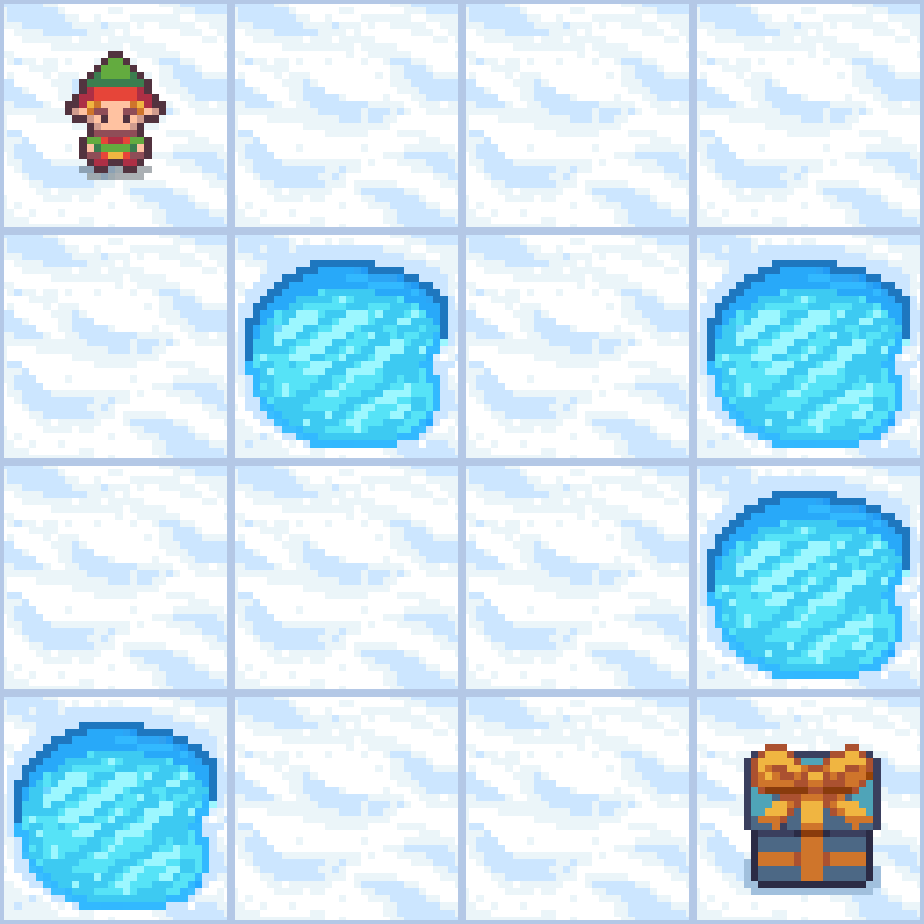}
            \caption{State}
            \label{fig:sub:gym4}
        \end{subfigure}
        \hfill
        \begin{subfigure}{0.49\linewidth}
            \centering
                \begin{tikzpicture}

        \foreach \x in {0,...,3} {
            \foreach \y in {0,...,3} {
              \draw[gray!60] (\x,\y) rectangle ++(1,1);
            }
        }

        \draw[black] (0, 0) rectangle ++(4, 4);
        
        \filldraw[fill=cb_green!30] (0,3) rectangle ++(1,1);
        \node at (0.5,3.5) {\textbf{S}};
        
        \filldraw[fill=cb_blue!30] (3,0) rectangle ++(1,1);
        \node at (3.5,0.5) {$g$};
        
        \foreach \x/\y in {
            1/2,
            3/2,
            0/0,
            3/1
        } {
            \filldraw[fill=cb_orange!40] (\x,\y) rectangle ++(1,1);
            \node at (\x+0.5,\y+0.5) {\textbf{H}};
        }
    
    \end{tikzpicture}
            \caption{Schema}
            \label{fig:sub:schem4}
        \end{subfigure}
    \end{subfigure}

    \caption{
        Schematic and image representations for the FrozenLake-v$1$ environment.
        In Figs.~\ref{fig:sub:schem4} and~\ref{fig:sub:schem8}, each cell corresponds to a Markovian state, and special tiles denote the start~(S), goal~(G), and terminal-failure states~(H).
        These structured state descriptions are encoded as contextual information and incorporated into prompts, supporting decision-making in the PPO policy through gated interventions.
    }
    \label{fig:frozenlake_schemas}
\end{figure}

We evaluate the environment under multiple grid sizes, including $4 \times 4$ and $8 \times 8$ configurations, as illustrated in Fig.~\ref{fig:frozenlake_schemas}.
For integration with language-model-based reasoning, the grid is additionally converted into a structured symbolic representation.
Each cell is encoded with its semantic type and used to construct local context descriptors around the agent position, including immediate neighboring tiles and limited look-ahead information.
These structured state descriptions are injected into prompts as contextual variables, enabling high-level reasoning about safety and goal proximity while preserving the underlying Markov decision process.

This formulation makes FrozenLake a controlled testbed for studying uncertainty-aware interventions, generalization, and the integration of RL policies with symbolic decision-making.

\section{Algorithm}

Algorithm~\ref{alg:method} outlines \abbrev.
Given a trained RL policy $\agent$, an LM $\generalist$, and a set of evaluation contexts $C' \in C$, where $C'$ are contexts not seen during training or during evaluation, \abbrev operates as follows.
For each context $c \in C'$, it initializes an environment instance (Alg.~\ref{alg:method}, Ln.~\ref{alg:ln:env}) and retrieves the initial observation $o$ (Alg.~\ref{alg:method}, Ln.~\ref{alg:ln:obs}).
At each step, the agent computes an action $a$ using its learned policy $\agent(o)$ (Alg.~\ref{alg:method}, Ln.~\ref{alg:ln:agent}).
However, before executing this action, \abbrev computes the total uncertainty $\mathcal{U}_e(o) + \mathcal{U}_a(o)$ using MC Dropout.
If the total uncertainty exceeds a predefined threshold $\threshold$ (Alg.~\ref{alg:method}, Ln.~\ref{alg:ln:uncertainty}), we override the agent's action by querying the LM with a prompt $Pr$ to obtain an alternative action $a' \gets \generalist(Pr)$ (Alg.~\ref{alg:method}, Ln.~\ref{alg:ln:slm}).
This intervention occurs only when the agent is uncertain, allowing the trained policy to handle familiar observations while delegating high-uncertainty decisions to the language model's world knowledge.
If the action generated by $\generalist$ is part of the set of valid actions $A$, then we overwrite the RL policy's action with the alternative one $a'$.
The agent then executes the selected action, receives the next observation from the environment via the transition function $T$, and repeats this process until it reaches the goal state $g$.
Finally, to select the threshold $\threshold$, we use Optuna~\cite{akiba2019optuna} and allow \abbrev to run on the evaluation set of contexts, which is OOD with respect to the test and train sets.
Because the evaluation set of contexts is fixed when selecting the RL policy and threshold, the benchmark remains fair across different model sizes and RL policies.
No optimization is done in the test set $C'$.

\begin{algorithm}[h!tbp]
    \footnotesize
    \caption{\method}
    \label{alg:method}
    \begin{algorithmic}[1]
        \Require A LM $\generalist$ and a prompt $Pr$ \label{alg:ln:require_init}
        \Require A trained RL policy $\agent$ and a threshold $\threshold$
        \Require An environment $Env$ and a set of Context $C' \in C$ \label{alg:ln:require_end}
        \For{each context $c \in C'$}
            \State Initialize environment instance $e$ with context $c$ \label{alg:ln:env}
            \State Retrieve first observation $o$ from $e$ \label{alg:ln:obs}
            \While{$\agent$ not at $g$} 
                \State $a \gets \agent(o)$ \label{alg:ln:agent}
                \If{$\mathcal{U}_e(o) + \mathcal{U}_a(o) \geqslant \threshold$} \label{alg:ln:uncertainty}
                    \State Create prompt $Pr$ with current observation $o$
                    \State $a' \gets \generalist(Pr)$ \label{alg:ln:slm}
                \EndIf
                \If{$a' \in A$}
                    \State $a \gets a'$
                \EndIf
                \State $o \gets T(o, a)$ \label{alg:ln:transition}
            \EndWhile
        \EndFor
    \end{algorithmic}
\end{algorithm}

\subsection*{Threshold Analysis}

The uncertainty threshold $\tau$ was selected independently for each model and environment 
size via Bayesian hyperparameter optimization using Optuna~\cite{akiba2019optuna}, searching over 
the continuous range $[0.10, 1.20]$ with 10 trials per study (22 studies in total). 
Table~\ref{tab:thresholds} reports the best $\tau$ found per model averaged across 
FrozenLake sizes $6{\times}6$, $7{\times}7$, and $8{\times}8$, alongside the corresponding 
average reward.

\begin{table}[h]
\centering
\caption{Best uncertainty threshold $\tau$ per model (averaged across environment sizes).}
\label{tab:thresholds}
\begin{tabular*}{.85\columnwidth}{l@{\extracolsep{\fill}}cc}
\toprule
\textbf{Model} & \textbf{Avg. $\tau$} & \textbf{Avg. Reward} \\
\midrule
Qwen2.5-0.5B  & 0.971 & 0.83 \\
Qwen2.5-1.5B  & 0.544 & 0.83 \\
Qwen2.5-3B    & 1.063 & 0.80 \\
Qwen2.5-7B    & 0.944 & 0.82 \\
Qwen2.5-14B   & 1.079 & 0.82 \\
Qwen2.5-32B   & 0.763 & 0.83 \\
Qwen2.5-72B   & 0.483 & 0.84 \\
\bottomrule
\end{tabular*}
\end{table}

The optimal thresholds reveal a pattern that mirrors the U-shaped performance curve 
discussed in Section~IV. Mid-size models (3B, 7B, and 14B) require high thresholds 
($\tau > 0.9$) to avoid degrading PPO performance, indicating that their guidance is 
unreliable and the gating mechanism must suppress LM intervention as much as possible. 
In contrast, the 1.5B model operates effectively with lower thresholds ($\tau \approx 0.54$), 
suggesting well-calibrated uncertainty estimates that allow more frequent, yet conservative, 
LM consultation. The 72B model achieves the lowest average threshold ($\tau = 0.48$) and 
the highest average reward, consistent with its ability to provide reliable guidance even 
when queried frequently — on $7{\times}7$, the optimal threshold drops to $\tau = 0.12$, 
meaning the LM is consulted at nearly every step, yet performance is maintained at 0.89. 
This contrast highlights that threshold sensitivity is tightly coupled with model capability: 
larger models tolerate lower thresholds because their action recommendations are trustworthy, 
while mid-size models require high thresholds to limit the damage from unreliable interventions.

\section{Evaluation Results}\label{ap:eval}
Tab~\ref{tab:ppo_lm_complete} displays the result for the same-size map task.
Tab.~\ref{tab:downward_generalization_eval_test} displays the results for the downwards generalization.
We note that the evaluation and the test PPO's performance remains approximate.
Therefore, the RL policy achieves results comparable to those obtained when it was selected during training.
Finally, we also note that all RL policies achieve the goal $100\%$, e.g., reward of $1.00 \pm 0.00$.

\begin{table*}[b]
\scriptsize
\centering
\caption{Comparison between PPO baseline and \abbrev on FrozenLake.
Eval corresponds to validation performance using the gating mechanism.
The test corresponds to in-domain test performance.
Results are reported as mean $\pm$ standard deviation over 100 episodes.}
\label{tab:ppo_lm_complete}
\begin{tabular*}{\textwidth}{lll @{\extracolsep{\fill}} cccc}
\toprule
Size & Model & Split & Reward & Length & IR (\%) & OR (\%) \\
\midrule

\multirow{16}{*}{$6\times6$}
 & \multirow{2}{*}{PPO} & Eval & $0.89 \pm 0.31$ & $9.22 \pm 2.32$ & -- & -- \\
 &                      & Test & $0.93 \pm 0.26$ & $9.49 \pm 1.89$ & -- & -- \\
 \cmidrule(lr){2-7}
 & \multirow{2}{*}{0.5B} & Eval & $0.89 \pm 0.31$ & $9.22 \pm 2.32$ & 21.10 & 0.02 \\
 &                       & Test & $0.93 \pm 0.26$ & $9.49 \pm 1.90$ & 49.83 & 0.00 \\
 \cmidrule(lr){2-7}
 & \multirow{2}{*}{1.5B} & Eval & $0.89 \pm 0.31$ & $9.22 \pm 2.32$ & 27.88 & 0.00 \\
 &                       & Test & $0.93 \pm 0.26$ & $9.49 \pm 1.90$ & 55.03 & 0.00 \\
 \cmidrule(lr){2-7}
 & \multirow{2}{*}{3B} & Eval & $0.74 \pm 0.44$ & $13.79 \pm 19.68$ & 25.32 & 8.32 \\
 &                     & Test & $0.93 \pm 0.26$ & $9.49 \pm 1.90$ & 0.00 & 0.00 \\
 \cmidrule(lr){2-7}
 & \multirow{2}{*}{7B} & Eval & $0.88 \pm 0.32$ & $10.14 \pm 7.92$ & 16.59 & 3.36 \\
 &                     & Test & $0.92 \pm 0.27$ & $11.66 \pm 12.95$ & 27.19 & 4.09 \\
 \cmidrule(lr){2-7}
 & \multirow{2}{*}{14B} & Eval & $0.75 \pm 0.43$ & $25.20 \pm 32.32$ & 32.30 & 17.84 \\
 &                      & Test & $0.92 \pm 0.27$ & $9.55 \pm 2.13$ & 0.97 & 0.47 \\
 \cmidrule(lr){2-7}
 & \multirow{2}{*}{32B} & Eval & $0.89 \pm 0.31$ & $9.22 \pm 2.32$ & 17.27 & 0.00 \\
 &                      & Test & $0.93 \pm 0.26$ & $9.49 \pm 1.90$ & 32.03 & 0.00 \\
 \cmidrule(lr){2-7}
 & \multirow{2}{*}{72B} & Eval & $0.89 \pm 0.32$ & $12.92 \pm 17.13$ & 19.37 & 5.30 \\
 &                      & Test & $0.93 \pm 0.26$ & $9.50 \pm 1.86$ & 24.03 & 0.60 \\
\midrule

\multirow{16}{*}{$7\times7$}
 & \multirow{2}{*}{PPO} & Eval & $0.86 \pm 0.36$ & $12.91 \pm 12.68$ & -- & -- \\
 &                      & Test & $0.86 \pm 0.35$ & $13.08 \pm 12.58$ & -- & -- \\
 \cmidrule(lr){2-7}
 & \multirow{2}{*}{0.5B} & Eval & $0.85 \pm 0.36$ & $12.73 \pm 11.60$ & 27.81 & 0.41 \\
 &                       & Test & $0.87 \pm 0.34$ & $11.45 \pm 2.46$ & 58.23 & 0.26 \\
 \cmidrule(lr){2-7}
 & \multirow{2}{*}{1.5B} & Eval & $0.85 \pm 0.36$ & $12.91 \pm 12.68$ & 30.60 & 0.00 \\
 &                       & Test & $0.86 \pm 0.35$ & $13.08 \pm 12.64$ & 41.59 & 0.00 \\
 \cmidrule(lr){2-7}
 & \multirow{2}{*}{3B} & Eval & $0.68 \pm 0.47$ & $23.56 \pm 31.05$ & 22.22 & 9.77 \\
 &                     & Test & $0.85 \pm 0.36$ & $13.03 \pm 12.65$ & 0.56 & 0.14 \\
 \cmidrule(lr){2-7}
 & \multirow{2}{*}{7B} & Eval & $0.81 \pm 0.39$ & $16.47 \pm 20.04$ & 21.16 & 6.13 \\
 &                     & Test & $0.86 \pm 0.35$ & $14.25 \pm 15.27$ & 9.31 & 2.53 \\
 \cmidrule(lr){2-7}
 & \multirow{2}{*}{14B} & Eval & $0.63 \pm 0.48$ & $35.63 \pm 38.79$ & 37.26 & 23.99 \\
 &                      & Test & $0.86 \pm 0.35$ & $13.08 \pm 12.64$ & 0.00 & 0.00 \\
 \cmidrule(lr){2-7}
 & \multirow{2}{*}{32B} & Eval & $0.85 \pm 0.36$ & $12.91 \pm 12.68$ & 23.84 & 0.00 \\
 &                      & Test & $0.86 \pm 0.35$ & $13.08 \pm 12.64$ & 56.62 & 0.00 \\
 \cmidrule(lr){2-7}
 & \multirow{2}{*}{72B} & Eval & $0.87 \pm 0.34$ & $12.99 \pm 11.39$ & 26.60 & 9.39 \\
 &                      & Test & $0.87 \pm 0.34$ & $12.22 \pm 9.09$ & 58.32 & 0.32 \\
\midrule

\multirow{16}{*}{$8\times8$}
 & \multirow{2}{*}{PPO} & Eval & $0.75 \pm 0.36$ & $12.01 \pm 12.68$ & -- & -- \\
 &                      & Test & $0.79 \pm 0.41$ & $12.49 \pm 3.17$ & -- & -- \\
 \cmidrule(lr){2-7}
 & \multirow{2}{*}{0.5B} & Eval & $0.75 \pm 0.43$ & $12.02 \pm 3.66$ & 24.16 & 0.05 \\
 &                       & Test & $0.79 \pm 0.41$ & $12.52 \pm 3.21$ & 59.18 & 0.13 \\
 \cmidrule(lr){2-7}
 & \multirow{2}{*}{1.5B} & Eval & $0.75 \pm 0.43$ & $12.01 \pm 3.66$ & 19.09 & 0.00 \\
 &                       & Test & $0.79 \pm 0.41$ & $12.49 \pm 3.19$ & 36.61 & 0.00 \\
 \cmidrule(lr){2-7}
 & \multirow{2}{*}{3B} & Eval & $0.48 \pm 0.50$ & $31.36 \pm 36.29$ & 42.33 & 22.48 \\
 &                     & Test & $0.71 \pm 0.46$ & $18.14 \pm 21.14$ & 14.74 & 6.63 \\
 \cmidrule(lr){2-7}
 & \multirow{2}{*}{7B} & Eval & $0.65 \pm 0.48$ & $22.89 \pm 27.77$ & 42.38 & 10.99 \\
 &                     & Test & $\mathbf{0.79 \pm 0.41}$ & $15.49 \pm 13.93$ & 14.99 & 2.88 \\
 \cmidrule(lr){2-7}
 & \multirow{2}{*}{14B} & Eval & $0.63 \pm 0.48$ & $26.52 \pm 31.56$ & 29.67 & 18.41 \\
 &                      & Test & $0.77 \pm 0.42$ & $15.10 \pm 15.10$ & 4.32 & 3.20 \\
 \cmidrule(lr){2-7}
 & \multirow{2}{*}{32B} & Eval & $0.75 \pm 0.43$ & $12.01 \pm 3.66$ & 25.07 & 0.00 \\
 &                      & Test & $0.79 \pm 0.41$ & $12.49 \pm 3.19$ & 43.01 & 0.00 \\
 \cmidrule(lr){2-7}
 & \multirow{2}{*}{72B} & Eval & $0.72 \pm 0.45$ & $19.07 \pm 22.77$ & 30.05 & 11.24 \\
 &                      & Test & $0.79 \pm 0.41$ & $12.50 \pm 3.17$ & 28.60 & 0.54 \\
\bottomrule
\end{tabular*}
\end{table*}

\begin{table*}[t]
\scriptsize
\centering
\caption{Downward generalization results. PPO trained on FrozenLake-8 and evaluated on smaller environments.
Eval corresponds to validation performance, while Test reports final performance using fixed thresholds.
Results are reported as mean $\pm$ standard deviation over 100 episodes.}
\label{tab:downward_generalization_eval_test}
\begin{tabular*}{\textwidth}{lll @{\extracolsep{\fill}} cccc}
\toprule
Env & LLM Model & Split & Reward & Length & IR (\%) & OR (\%) \\
\midrule

\multirow{14}{*}{$4\!\times\!4$}
 & \multirow{2}{*}{0.5B} & Eval & $0.00 \pm 0.00$ & $51.76 \pm 46.85$ & 100.00 & 57.13 \\
 &                       & Test & $0.00 \pm 0.00$ & $41.12 \pm 26.20$ & 100.00 & 61.71 \\
 \cmidrule(lr){2-7}
 & \multirow{2}{*}{1.5B} & Eval & $0.47 \pm 0.50$ & $7.54 \pm 6.46$ & 100.00 & 41.00 \\
 &                       & Test & $0.37 \pm 0.49$ & $6.92 \pm 6.20$ & 100.00 & 42.98 \\
 \cmidrule(lr){2-7}
 & \multirow{2}{*}{3B}   & Eval & $0.00 \pm 0.00$ & $83.40 \pm 36.87$ & 100.00 & 62.20 \\
 &                       & Test & $0.00 \pm 0.00$ & $44.58 \pm 26.47$ & 100.00 & 64.52 \\
 \cmidrule(lr){2-7}
 & \multirow{2}{*}{7B}   & Eval & $0.00 \pm 0.00$ & $100.00 \pm 0.00$ & 100.00 & 98.70 \\
 &                       & Test & $0.00 \pm 0.00$ & $61.00 \pm 0.00$ & 100.00 & 95.57 \\
 \cmidrule(lr){2-7}
 & \multirow{2}{*}{14B}  & Eval & $0.03 \pm 0.17$ & $95.22 \pm 20.95$ & 100.00 & 55.80 \\
 &                       & Test & $0.07 \pm 0.26$ & $56.59 \pm 15.04$ & 100.00 & 52.95 \\
 \cmidrule(lr){2-7}
 & \multirow{2}{*}{32B}  & Eval & $0.93 \pm 0.26$ & $9.68 \pm 18.55$ & 100.00 & 50.66 \\
 &                       & Test & $0.95 \pm 0.22$ & $7.70 \pm 9.46$ & 100.00 & 46.17 \\
 \cmidrule(lr){2-7}
 & \multirow{2}{*}{72B}  & Eval & $0.94 \pm 0.24$ & $9.98 \pm 18.50$ & 100.00 & 68.98 \\
 &                       & Test & $0.95 \pm 0.22$ & $7.99 \pm 9.51$ & 100.00 & 64.22 \\
\midrule

\multirow{14}{*}{$5\!\times\!5$}
 & \multirow{2}{*}{0.5B} & Eval & $0.01 \pm 0.10$ & $29.21 \pm 36.82$ & 86.76 & 86.75 \\
 &                       & Test & $0.00 \pm 0.00$ & $31.61 \pm 27.21$ & 100.00 & 99.87 \\
 \cmidrule(lr){2-7}
 & \multirow{2}{*}{1.5B} & Eval & $0.23 \pm 0.42$ & $10.10 \pm 10.55$ & 100.00 & 32.43 \\
 &                       & Test & $0.23 \pm 0.42$ & $11.53 \pm 13.13$ & 100.00 & 28.47 \\
 \cmidrule(lr){2-7}
 & \multirow{2}{*}{3B}   & Eval & $0.00 \pm 0.00$ & $79.04 \pm 40.00$ & 89.64 & 39.77 \\
 &                       & Test & $0.00 \pm 0.00$ & $50.92 \pm 22.39$ & 100.00 & 41.07 \\
 \cmidrule(lr){2-7}
 & \multirow{2}{*}{7B}   & Eval & $0.00 \pm 0.00$ & $92.20 \pm 26.59$ & 88.44 & 46.62 \\
 &                       & Test & $0.00 \pm 0.00$ & $61.00 \pm 0.00$ & 100.00 & 51.03 \\
 \cmidrule(lr){2-7}
 & \multirow{2}{*}{14B}  & Eval & $0.01 \pm 0.10$ & $94.73 \pm 21.28$ & 87.34 & 48.86 \\
 &                       & Test & $0.01 \pm 0.10$ & $58.11 \pm 12.70$ & 100.00 & 47.27 \\
 \cmidrule(lr){2-7}
 & \multirow{2}{*}{32B}  & Eval & $0.85 \pm 0.36$ & $18.34 \pm 28.91$ & 100.00 & 49.51 \\
 &                       & Test & $0.87 \pm 0.34$ & $10.12 \pm 10.71$ & 100.00 & 47.55 \\
 \cmidrule(lr){2-7}
 & \multirow{2}{*}{72B}  & Eval & $0.85 \pm 0.36$ & $22.54 \pm 33.10$ & 100.00 & 51.05 \\
 &                       & Test & $0.86 \pm 0.35$ & $16.18 \pm 18.42$ & 100.00 & 50.51 \\
\midrule

\multirow{14}{*}{$6\!\times\!6$}
 & \multirow{2}{*}{0.5B} & Eval & $0.06 \pm 0.24$ & $16.83 \pm 29.10$ & 61.68 & 61.68 \\
 &                       & Test & $0.00 \pm 0.00$ & $31.61 \pm 25.91$ & 100.00 & 100.00 \\
 \cmidrule(lr){2-7}
 & \multirow{2}{*}{1.5B} & Eval & $0.11 \pm 0.31$ & $9.76 \pm 9.79$ & 100.00 & 30.39 \\
 &                       & Test & $0.11 \pm 0.31$ & $8.60 \pm 8.27$ & 100.00 & 28.96 \\
 \cmidrule(lr){2-7}
 & \multirow{2}{*}{3B}   & Eval & $0.00 \pm 0.00$ & $80.63 \pm 38.97$ & 99.99 & 41.64 \\
 &                       & Test & $0.00 \pm 0.00$ & $50.38 \pm 22.78$ & 100.00 & 41.43 \\
 \cmidrule(lr){2-7}
 & \multirow{2}{*}{7B}   & Eval & $0.00 \pm 0.00$ & $38.16 \pm 46.70$ & 9.94 & 4.90 \\
 &                       & Test & $0.00 \pm 0.00$ & $61.00 \pm 0.00$ & 100.00 & 50.31 \\
 \cmidrule(lr){2-7}
 & \multirow{2}{*}{14B}  & Eval & $0.00 \pm 0.00$ & $100.00 \pm 0.00$ & 100.00 & 51.88 \\
 &                       & Test & $0.01 \pm 0.10$ & $60.31 \pm 5.17$ & 100.00 & 51.50 \\
 \cmidrule(lr){2-7}
 & \multirow{2}{*}{32B}  & Eval & $0.70 \pm 0.46$ & $31.15 \pm 37.31$ & 100.00 & 53.56 \\
 &                       & Test & $0.69 \pm 0.46$ & $22.92 \pm 21.88$ & 100.00 & 52.35 \\
 \cmidrule(lr){2-7}
 & \multirow{2}{*}{72B}  & Eval & $0.74 \pm 0.44$ & $29.05 \pm 35.95$ & 100.00 & 50.74 \\
 &                       & Test & $0.75 \pm 0.44$ & $21.58 \pm 21.09$ & 100.00 & 50.33 \\
\midrule

\multirow{14}{*}{$7\!\times\!7$}
 & \multirow{2}{*}{0.5B} & Eval & $0.00 \pm 0.00$ & $28.31 \pm 42.73$ & 0.00 & 0.00 \\
 &                       & Test & $0.00 \pm 0.00$ & $25.28 \pm 24.98$ & 100.00 & 99.95 \\
 \cmidrule(lr){2-7}
 & \multirow{2}{*}{1.5B} & Eval & $0.08 \pm 0.27$ & $8.18 \pm 7.67$ & 100.00 & 28.84 \\
 &                       & Test & $0.10 \pm 0.30$ & $8.08 \pm 7.06$ & 100.00 & 29.14 \\
 \cmidrule(lr){2-7}
 & \multirow{2}{*}{3B}   & Eval & $0.00 \pm 0.00$ & $28.31 \pm 42.73$ & 0.00 & 0.00 \\
 &                       & Test & $0.00 \pm 0.00$ & $53.51 \pm 19.53$ & 100.00 & 44.23 \\
 \cmidrule(lr){2-7}
 & \multirow{2}{*}{7B}   & Eval & $0.00 \pm 0.00$ & $28.31 \pm 42.73$ & 0.00 & 0.00 \\
 &                       & Test & $0.00 \pm 0.00$ & $61.00 \pm 0.00$ & 100.00 & 51.57 \\
 \cmidrule(lr){2-7}
 & \multirow{2}{*}{14B}  & Eval & $0.00 \pm 0.00$ & $28.31 \pm 42.73$ & 0.00 & 0.00 \\
 &                       & Test & $0.00 \pm 0.00$ & $61.00 \pm 0.00$ & 100.00 & 50.51 \\
 \cmidrule(lr){2-7}
 & \multirow{2}{*}{32B}  & Eval & $0.00 \pm 0.00$ & $28.31 \pm 42.73$ & 0.00 & 0.00 \\
 &                       & Test & $0.58 \pm 0.50$ & $30.20 \pm 23.38$ & 100.00 & 53.36 \\
 \cmidrule(lr){2-7}
 & \multirow{2}{*}{72B}  & Eval & $0.68 \pm 0.47$ & $36.76 \pm 38.98$ & 100.00 & 52.18 \\
 &                       & Test & $0.68 \pm 0.47$ & $27.10 \pm 21.94$ & 100.00 & 50.88 \\
\bottomrule
\end{tabular*}
\end{table*}

\end{document}